\documentclass{article}

\usepackage[final]{corl_2019} 

\usepackage{amsmath}
\usepackage{graphicx}
\usepackage{subfigure} 
\DeclareMathOperator{\argmin}{argmin}
\title{EnforceNet: Monocular Camera Localization in Large Scale Indoor Sparse LiDAR Point Cloud}

\author{
  Yu Chen\\
  Blacksesame Technologies \\
  \texttt{yu.chen@bst.ai} \\
  \And
  Guan Wang \\
  Blacksesame Technologies \\
  \texttt{guan.wang@bst.ai} \\
}

\begin{document}
\maketitle


\begin{abstract}
Pose estimation is a fundamental building block for robotic applications such as autonomous  vehicles, UAV, and large scale augmented reality. It is also a prohibitive factor for those applications to be in mass production, since the state-of-the-art, centimeter-level pose estimation often requires long mapping procedures and expensive localization sensors, e.g. LiDAR and high precision GPS/IMU, etc. To overcome the cost barrier, we propose a neural network based solution to localize a consumer degree RGB camera within a prior sparse LiDAR map with comparable centimeter-level precision. We achieved it by introducing a novel network module, which we call ``resistor module'', to enforce the network generalize better, predicts more accurately, and converge faster. Such results are benchmarked by several datasets we collected in the large scale indoor parking garage scenes. We plan to open both the data and the code for the community to join the effort to advance this field.
\end{abstract}



\section{Introduction}
For a robot to navigate in space autonomously, it has to localize itself precisely within a prior map of the environment. Therefore, 6DoF pose estimation of the localization sensor is one of the core technologies to enable robots such as autonomous cars into reality. Nowadays, the most commonly used localization sensors are camera and LiDAR. LiDAR generates the point cloud of the environment as the map, and the localization system would find the best registration between the runtime point cloud w.r.t a subarea of the map to infer the LiDAR pose~\citep{sebastian07}. With a good guess of initial position (often from a high definition GPS), this approach infers the pose at the centimeter level, the start-of-the-art accuracy for self-driving cars. However, there are several roadblocks for this approach to be in mass production. First, the cost of LiDAR and HD GPS remains too high for consumers. Second, the engineering complexity of a LiDAR also challenges the manufacturers to produce reliable items. Third, the GPS signal sometimes is simply not available, such as the indoor scenario, e.g., parking garages, etc. Without a good initial pose, the LiDAR localization system would require significantly increased computation resources at least, if not fail completely. The above limitations of LiDAR-based localization approach have been motivating camera-based approaches. Cameras are cheap in price, easy to mass produce, and, thanks to the advancement of visual odometry (VO)\citep{forster2014svo} \citep{engel2017direct} and visual SLAM (vSLAM)~\citep{davison2007monoslam} \citep{klein2009parallel} \citep{engel2014lsd} \citep{mur2017orb}, camera pose could be precisely computed given only visual inputs, without GPS or with consumer degree GPS. However, the achilles heel for camera-based localization is its stability against different lighting conditions and the structure of the visual scenes, and its inaccurate perception of the scene depth~\citep{cadena2016past}. When the robot movement creates a strong uneven lighting sequence, e.g., sudden exposure of strong light from dark shadow, state-of-the-art direct VO or vSLAM will inevitably fail, i.e., loss tracking or fluctuate too significantly to be useful. Even photometric calibration~\citep{bergmann2017online} would only help in a scratching degree instead of solving the problem. Furthermore, visual methods often rely on the presence of structures at their scene, so the algorithms may find many ``features'' to track across frames. Although such a requirement is often fulfilled in outdoor environments, it could be lacking in indoor scenes. For example, in typical parking garages, large white walls and repetitive pillars often occupy the images, making the indirect VO/vSLAM methods that leveraging geometry features suffer. Last but not least, when accurate depth of the scene is not available, the visual methods may result in scale drift~\citep{cadena2016past} that leads to sizable accumulative localization error within a few hundred frames, which makes it not applicable for a reasonable trip length (thousands of frames). 

There emerges a handful of initiatives that attempt to combine the advantages from both the LiDAR side and the camera side. That is to estimate camera pose using its image inputs within a prior 3D LiDAR point cloud map. The motivations along this line of research are clearly attractive. First, despite the cost of LiDAR, using it for survey purposes of the environment has been a feasible industrial approach for decades. The result point cloud is the most accurate we can get based on current technology, overshadowing mapping methods by cameras in terms of stability and accuracy~\citep{Geiger2012CVPR}. Second, cameras capture the semantics of the world orders of magnitude more than LiDAR, and they are more accessible than LiDAR on the end products that contain the localization function at the edge. However, the challenges of such research are also hard to overcome. To begin with, LiDAR and camera belong to different regimes in terms of sensing capabilities. Major localization approaches only operate in one regime. For example, LiDAR SLAM compares point cloud structures, and vSLAM matches image features. How to bridge ideas that work in one modal within another modal remains an active field of research. Secondly, the design space for the camera on vSLAM is huge. Monocular, stereo and RGB-d cameras are often referred to as major branches for the hardware used in vSLAM~\citep{cadena2016past}. Each of them brings unique leverages and constraints for algorithm designs. Which combination of a special camera setting and LiDAR map would be optimal is still an unanswered question. Thirdly, since LiDAR scans are often sparse and it is time-consuming, expensive, and technical challenging to obtain dense results, it is highly desirable to make the camera localization work within the sparse point cloud. Meanwhile, the density is an indicator of the quantity of information withheld in the point cloud. How well the camera can compensate the point cloud sparsity is another interesting challenge to face.  

In this paper, we propose an end-to-end novel neural network structure as a solution in pursuit of camera pose estimation in LiDAR point cloud that copes with all challenges above. We design the approach with a monocular camera which is the most commoditized among other types of camera in terms of hardware maturity and calibration difficulty. Given a mono camera image and initial rough pose estimation, our approach first makes a set of depth projections using the point cloud and feeds the pair of image and projection to infer a camera pose. More importantly, borrowing from the reinforcement learning framework, we designed a state value constraint, which we call ``resistor module'', in the network, to quantify how good the pose estimation could be, and backpropagates that supervision to the network. We found that the resistor module not only makes the network converges faster but also give it better predict power and generalization capability. 

To summarize, our contributions are as follows. 
\begin{itemize}
    \item We give a systematic solution for localizing camera within a sparse 3D LiDAR point cloud. Our solution has the level of accuracy that is on par with the state-of-the-art LiDAR-based localization method. Our output pose is an absolute scale, which is not normalized. Since it is based on end-to-end neural network training and inference, the system is also easy to implement and extend. Our designated model is a good trade-off accuracy and speed.
    \item With our approach to the localization problem, we invented a novel resistor module based on the idea of reinforcement learning, SARSA~\citep{sutton2018reinforcement} scheme to be specific. The new module significantly improved the network performance in multiple dimensions based on our evaluation results. 
    \item To the best of our knowledge, we are the first to conduct throughout benchmark on multi-modal localization problem in a large scale indoor setting. We build the agent for rendering the depth image for further data collection and augmentation.
    \item We will open-source our localization approach along with the data we have collected. In our future work, we are looking to extend such open framework to an open benchmarking tool for both indoor and outdoor localization research. 
\end{itemize}


\section{Related Work}
\label{sec:citations}


\textbf{Camera localization in 3D LiDAR Map.} Localization with cameras in 3D LiDAR map has been an active field in the autonomous driving like a typical sensor fusion direction. The major types of cameras used for this purpose are monocular, RGB-d, and stereo cameras. With different kinds of cameras, there are two large categories of methods to get the relative pose of the camera in 3D prior map. The first category is utilizing current view feature point to match the prior map points, the methodology behind this is to minimize the matching points' distance ~\citep{caselitz2016monocular,caselitz2016matching,gawel2016structure,saurer2016image}. This kind of methods need a good initial guess of scale if the monocular camera is used, and it suffers scale drift. For RGB-d or stereo cameras, the sensor cost will be much higher and the calibration process will be tedious. Besides, the LiDAR map points may not be the feature points in the camera image, and the correspondences between the LiDAR map and the camera image will be quite sparse, especially with the LiDAR of low beam lines. To avoid those constraints, our method belongs to the second category which renders synthetic views from the 3D map and comparing them with RGB images. With different information in the maps, the rendering can be different, like depth image, intensity image or RGB image \citep{pandey2015automatic,napier2013cross,wolcott2014visual,pascoe2015direct,neubert2017sampling}. Different criteria \citep{pandey2015automatic,pascoe2015direct,caron2014direct}  are proposed to measure the similarity of the camera image and the rendered image. We have found that this kind of method is suitable when the LiDAR map is dense and the pose transformation between the current camera view and the rendering image is not too large. However, with the changing light condition and scale drift, the method still suffers to pose jump and accumulative localization error.
There are earlier attempts that learn the similarity measurements with data instead of handcrafting them \citep{naseer2017deep,wang2018dels,radwan2018vlocnet++,wang2019densefusion}. These methods combine many tasks, such as semantic segmentation, depth prediction, or pose estimation, into one network and manipulate its inference heads and losses to learn them all at once. Our method is more focus on a single task and we innovate on the communication of the state-value prediction and the pose regression, we also design the network for high inference speed.

\textbf{Pose estimation.} Camera pose estimation is also an essential task in visual odometry or full-fledged vSLAM. Traditionally, these methods focus on minimizing the geometry error~\citep{mur2017orb} of the features or the photometric error \citep{engel2017direct} of the consecutive pixels. And the bundle adjustment \citep{triggs1999bundle} is usually used in the backend optimization procedure. Recently, the learning-based methods for pose estimation is becoming a hot topic, since they could overcome the motion blur or lighting problems to some extent \citep{kendall2015posenet,zhou2017unsupervised,li2018undeepvo,mahjourian2018unsupervised,casser2019unsupervised}. Most of learning based methods extract the relationship between consecutive frames with the photometric loss optimization. They do not study multi-modal setting like ours, and we do not assume consecutive frames as input. Therefore, our method is more suitable for camera re-localization, which complements VO or vSLAM.

\textbf{Deep reinforcement learning} Deep RL was applied to localization at a coarse degree~\citep{mirowski2018learning}. We have different problem settings and we did not discretize action space. Moreover, to avoid the complexity in training continuous RL agent, we focus more on the state-value only. 

	

\section{Dataset}
\begin{figure}
    \centering
    \includegraphics[width=0.3\textwidth]{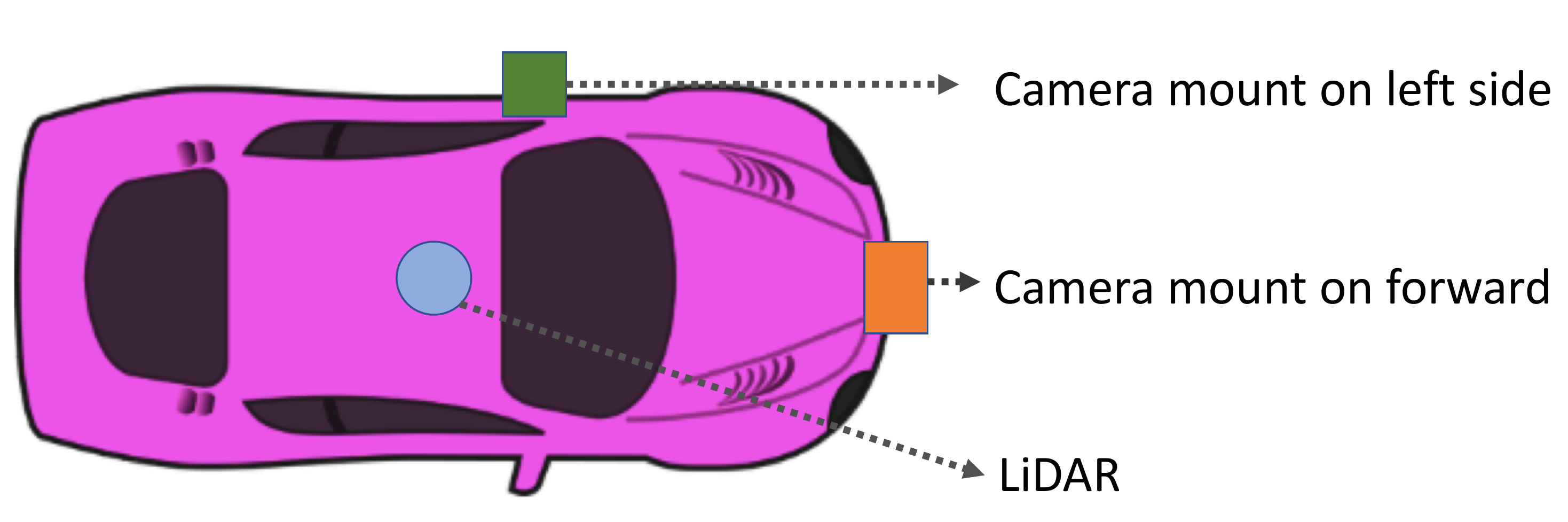}
    \caption{Data collection vehicle illustration}
    \label{fig:equipment}
\end{figure}

\textbf{Data collection} As we are focusing on the large scale closed area low-speed autonomous driving, but there are no such public open datasets with the 3D point cloud, RGB images and relative poses. We setup our own device for data collection. We used VLP-16 Velodyne LiDAR, and one global shutter Intel RealSense camera is fused with it by Autoware \citep{kato2018autoware} camera and LiDAR calibration toolkit. Figure~\ref{fig:equipment} illustrates our data collection vehicle. We modified the LeGO-LOAM SLAM \citep{shan2018lego} to construct the 3D point cloud for our data collection requirement. In total, we have about 30GB LiDAR and camera data for 100K frames at several different parking garages in California, US. We define $G_{ijk}$ as the $j^{th}$ trajectory data in the $i^{th}$ garage with camera mounting position $k$ ($k=0$ for forward facing and $k=1$ for side facing).



\section{Method}
\label{sec:method}
Intuitively, we design the approach by comparing the camera image with a set of depth images that are projected from several pose guesses, and quantifies the proximity of each guess towards the real camera pose to decide the sample positions for the next iteration of pose guesses. 
Therefore, our approach operates on image feature space rather than inferring depth map from images and operate on point cloud space. The first reason is that we can make an infinite number of accurate depth projections for any feasible camera parameters given a point cloud. The second reason is that image similarity measurement is a more diverse research field than point cloud registration, which means we may tackle the problem at more angles. More details about the approach are in the following subsections. 

\subsection{Problem Formulation}
Formally, a 6DoF camera pose is defined as $p = [R, t] \in SE(3)$ where $R \in SO(3)$ is the rotation and $t \in R^3$ is the translation. A LiDAR map is a set of points $P(L)$ from space $M = \{m_i | m_i \in R^3, i = 1, 2, 3, ... |M|\}$. A projection $P$ of a LiDAR map from the viewpoint pose $V_p$ is defined as 
\begin{equation}
    P(V_p)=G\cdot K\cdot \begin{bmatrix}
R_{V_p} & t_{V_p}\\ 
0 & 1
\end{bmatrix} P(L)
\end{equation}
Here $G$
is the projection parameter for clipping planes, and 
$K$ is the camera intrinsic parameters matrix. 

A pair of RGB image $I$ (taken at pose $P_i$) and a depth map $D$ (projected at pose $P_d$) is defined as \begin{equation}
    H_{P_iP_d} = I_{P_i} \otimes D_{P_d}
\end{equation}, where the $\otimes$ operator computes a representation vector for the pair. The pose difference of $I_{P_i}$ and $D_{P_d}$ is defined as \begin{equation}
    \Delta P_{i,d} = P_{i} - P_{d}, \Delta P_{i,d} \in SE(3)
\end{equation} 

A pose estimation problem can be therefore formulated as 
\begin{equation}
     \widetilde{\Delta P} = \underset{P_d}{\argmin} E(H_{P_i, P_d})
\end{equation}
 where $E$ is some function that quantifies the ``similarity'' between $I_{P_i}$ and $D_{P_d}$. For the ease of presentation, we will refer the pose used for depth projection as $P_d$ and the pose from which the image is taken as $P_i$. 

\begin{figure}
    \centering
    \includegraphics[width=0.25\textwidth]{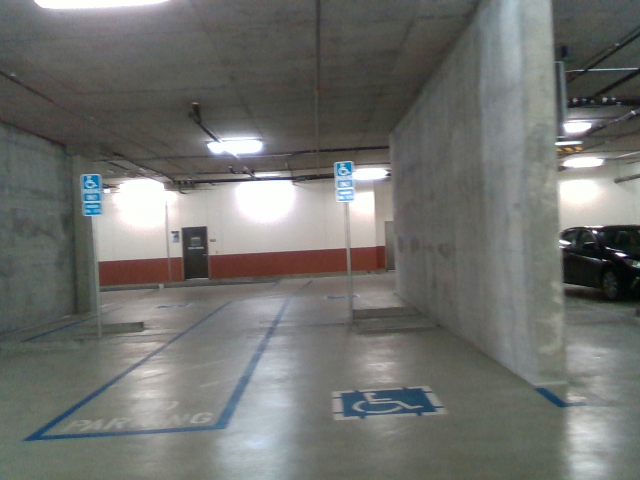}
    \hfill
    \includegraphics[width=0.25\textwidth]{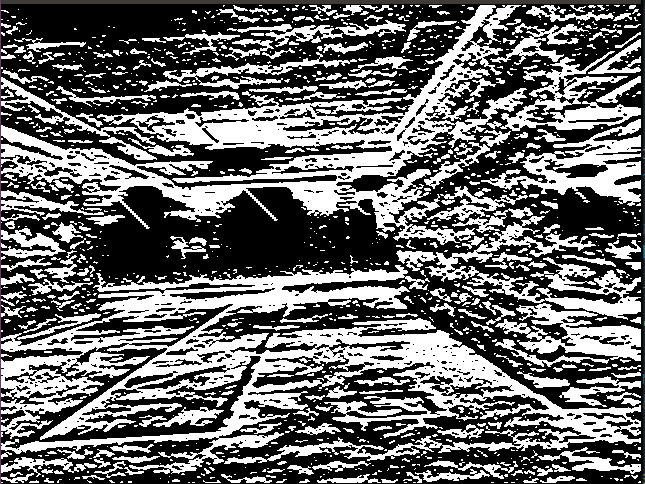}
    \hfill
    \includegraphics[width=0.25\textwidth]{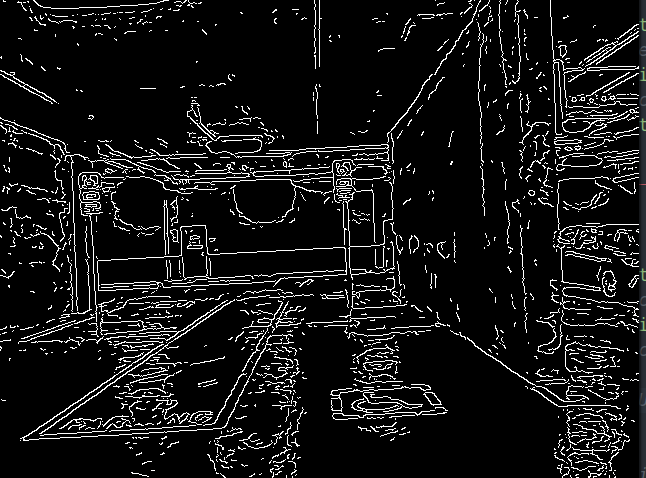}
    \includegraphics[width=0.25\textwidth]{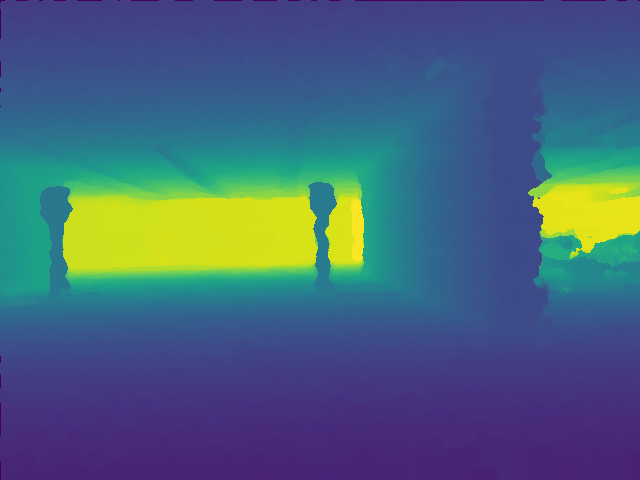}
    \hfill
    \includegraphics[width=0.25\textwidth]{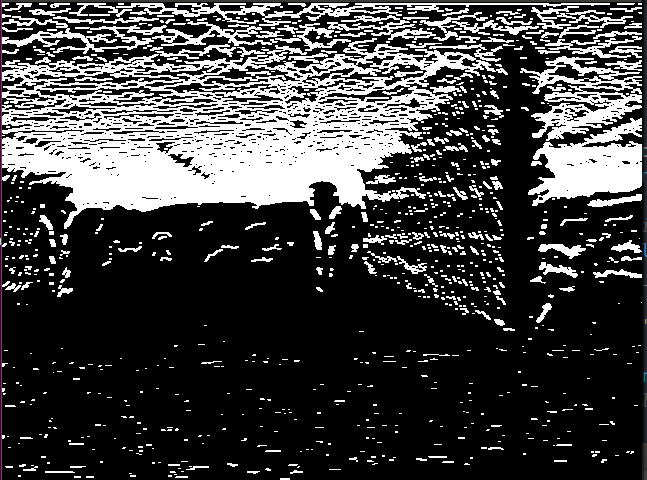}
    \hfill
    \includegraphics[width=0.25\textwidth]{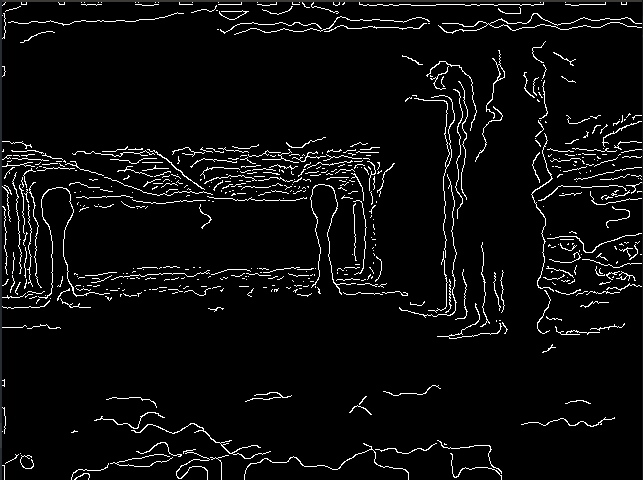}
    \caption{Example of RGB and Depth Map. The upper 3 images are RGB image and relative processing images. The bottom 3 images are depth image and relative processing images.}    
    \label{fig:rgb_map}
\end{figure}
 

\subsection{Exploration on Projective Image Registration}
In previous research~\citep{wolcott2014visual,neubert2017sampling}, $I_{P_i}$ and $D_{P_d}$ are brought to the same grey-scale or edge imagery landscape for a direct comparison by some mutual information based similarity measurement $E_{entropy}$. It has been demonstrated that if the ground truth $ \Delta P_{i,d}$ is small, their gradients trend to be close~\citep{neubert2017sampling}. 
However, we found that the gradient was too noisy to be useful if the depth is sparse. Figure~\ref{fig:rgb_map} shows such an example. The depth map is projected within a 3D point cloud generated by a Velodyne 16 line LiDAR one-time scan. Therefore, its gradient was far from smooth even after we applied a sequence of filters such as Gaussian or local contrast normalization.

We also tried to apply depth completion and triangulation to $D_{P_d}$, combined with different gradient filters such as Sobel or Canny. But from Figure~\ref{fig:rgb_map} shows no promising similar features on generating meaningful structures that could possibly match the sketches in $I_{P_i}$, as the depth image and the RGB image have different feature response from different sensors and lack of consistency in features.

We further confirmed it by applying several forms of $E_{entropy}$~\citep{wolcott2014visual,neubert2017sampling} and little matches were found between $I$ and $D$ taken at the same pose. 

This experiment confirmed that, due to sparsity, the mutual information is not significant enough to be quantified between $I_{P_i}$ and $D_{P_d}$ even if they are taken at the same pose. Therefore, the heuristic post-processing on the sparse depth map and the similarity measurement does not work well, which makes us leaning towards learning-based approaches. 

\subsection{EnforceNet Design}

\textbf{Goals and Specialties} Since the camera pose estimation is often a real-time task that requires reliable results given various lighting conditions, our design goals for the network 
are fast, accurate, and generalizable. Our scenario is low-speed large scale indoor and our algorithm should not depend on the sequence of frames, only the current image frame is utilized. Ideally, it should run at high frequency on embedded hardware with restricted computation resources. The localization error should be at centimeter-level. It needs to require as less retraining/fine-tuning as possible when applying to different scenes. It should be stable when facing lighting condition changes. 

To fulfill the goals, we fully leverage the prior 3D point cloud $M$ and the fact that infinite depth projections $\{D_j | j \in [1, \infty)\}$ can be generated from it with poses samples $\{P_d | d \in [1, \infty)\}$. This would help the model with not only exploration of training pairs of $H_{i,d}$ with large or small pose differences, but also exploitation of incrementally deriving plausible poses that are close to $P_i$.

\textbf{Resistor Module} In order to infer $P_i$ for one RGB image, we can randomly sample a set of $P_d$ within $M$ and measure $E(H_{P_i, P_d})$ to derive where our next batch of $\{P_d\}$ should be until they converge. This process can be formulated as a $MDP$ where the future pose estimates are independent of the past given the present estimates, of which the reinforcement learning frameworks may be suitable for solving. However, being a continuous-valued regression problem, it may be hard for the off-the-shelf RL frameworks, such as DQN, to tackle. It usually takes careful discretization of the action space~\citep{mirowski2018learning}, tremendously large dataset, and a large total number of episodes to successfully train a continuous RL model. Due to resource limitations, we take the inspiration from the RL frameworks and redirect it into a neural network design. 

\begin{figure}[!bp]
  \centering
  \begin{minipage}[b]{0.55\textwidth}
    \includegraphics[width=\textwidth]{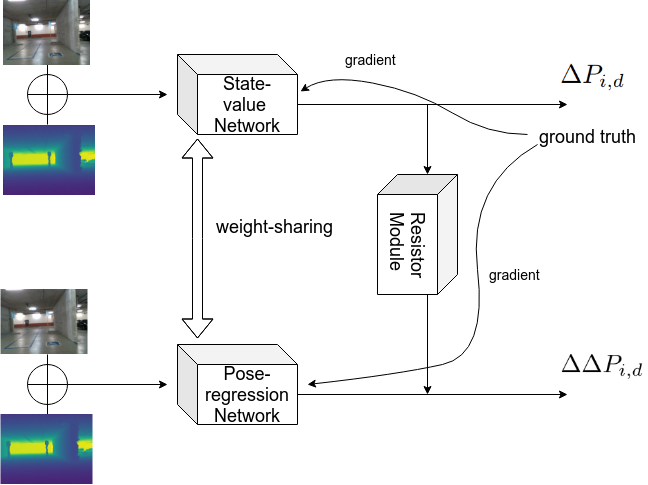}
    \caption{EnforceNet Sketch}
    \label{fig:state_value_net}
  \end{minipage}
  \hfill
  \begin{minipage}[b]{0.44\textwidth}
    \includegraphics[width=\textwidth]{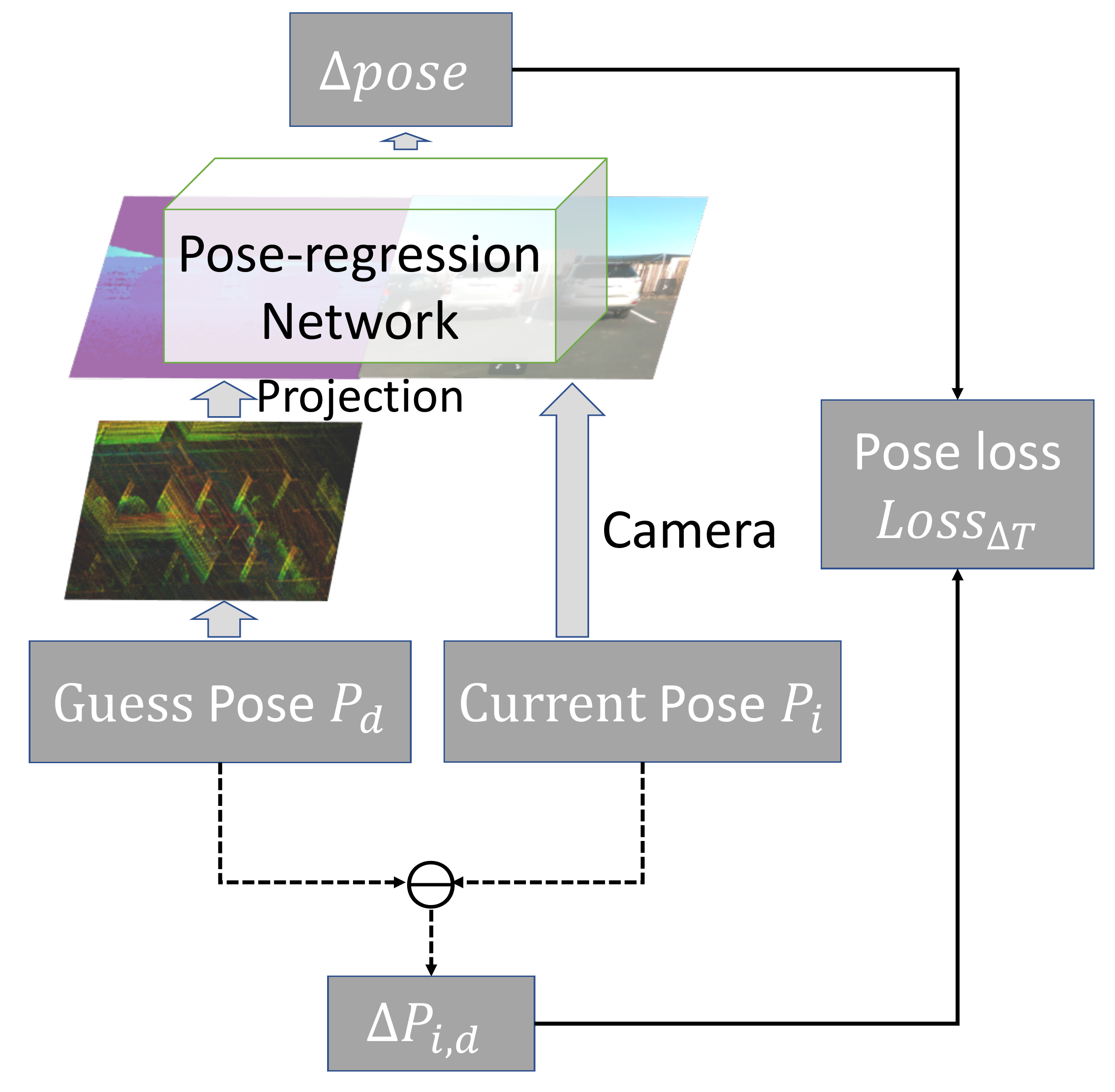}
    \caption{Pose-regression Network}
    \label{fig:pose_regress_net}
  \end{minipage}
\end{figure}

Implicitly, we have a virtual ``agent'' that explores the space for the best pose. We treat $H_{P_i, P_d}$ as a ``state'' for that agent. We define a state-value function $F(\Delta P_{i, d})$ to be a monotone decreasing function. Therefore, a state has a high value if $\Delta P_{i,d}$ is low, and low-value vice versa. In our context, $F$ is a convolutional neural network. In a classic SARSA paradigm, the state value function is helpful in selecting good actions which lead to better rewards and better-valued states. Since in the scope of this paper we neither discretize actions nor explicitly design reward function, we emphasize on discovering how the state-value function introduces regularization to the network, which makes it performs better in convergence, accuracy, and generalization ability. In addition to a neural network that quantifies state-value, we need another network to regress $\Delta P_{i,d}$ showing in Figure~\ref{fig:pose_regress_net}. The input for this network is also the RGB-depth image pairs $H_{P_i, P_d}$, and its labels are their ground truth pose difference. 


The regression network and the state-value network are connected by the resistor module, where the $\Delta P_{i,d}$ predictions are used as state-value labels. The intuition is that if the $\Delta P_{i,d}$ is large, the state value should be small, since the depth image should be from a projection that is far from the RGB image, while the $\Delta pose$ being small means the state value should be large. Therefore, the resistor module enforces the state-value network to learn about the ground truth $\Delta P_{i,d}$. Meanwhile, we decide to let both networks to share weights. The weight-sharing further enforces several benefits in terms of learning. First, it adds regulation on the $\Delta P_{i,d}$ regression network, because the state-value gradient with the ground truth $\Delta P_{i,d}$ also back-propagates through the regression network. Second, the state-value network learns the ground truth information easier, because the regression weights directly come from that. Third, the training complexity and the inference speed are both improved. Similar weight sharing scheme and the benefits are also discovered in AutoML research~\citep{pham2018efficient}. Figure~\ref{fig:state_value_net} shows how the whole architecture looks like. 

\begin{figure}
    \centering
    \includegraphics[width=0.85\textwidth]{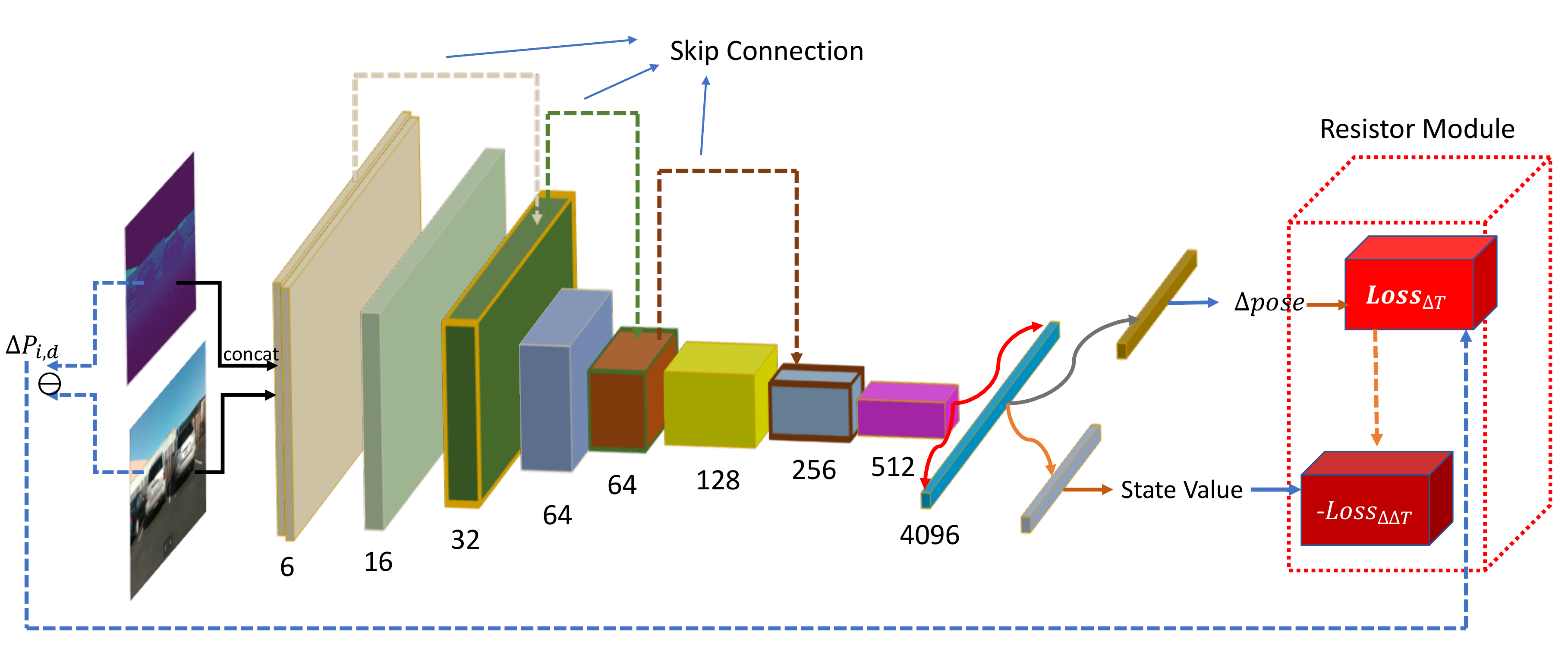}
    \caption{EnforceNet}
    \label{fig:enforcenet}
\end{figure}

\textbf{Network Architecture and Loss} We chose the backbone network to be a 7-layer ConvNet as depicted in Figure~\ref{fig:enforcenet}. The RGB and depth image pair $H_{P_i, P_d}$ is the network input. The $\otimes$ operator is stacking in this particular case. Their pose difference $\Delta P_{i,d}$  is the label. We applied some standard architectural tricks to the network, such as batch normalization and residual module. We used RMSProp as the back-propagation scheme. In contrast to more complex architectures in Figure~\ref{fig:enforcenet} that infers not only camera pose but also semantic scene or depth, our model is lightweight and more accurate on the localization task.

To train the network, the simple Euclidean error of combining translation and rotation is utilized. The ground truth transformation $\Delta P_{i,d}$ (translation $t_t$ and rotation $R_t$) at timestamp $t$ between the depth image and the camera capture can be described as: 

\begin{equation}
{\Delta P_{t,i,d}} = [ \Delta\textbf{R}_{t,i,d},\Delta\textbf{t}_{t,i,d}]
\end{equation}

The prediction transformation $\widetilde{\Delta P}$ can be combined with:

\begin{equation}
\widetilde{\Delta P} = [ \widetilde{\Delta\textbf{R}_{t}}, \widetilde{\Delta\textbf{t}_{t}}]
\end{equation}

The pose loss can be defined as the weighted sum of error of the two components from rotation and translation below ($\alpha_1$ is the rotation loss in the pose loss, $\alpha_2$ is the translation loss in the pose loss):

\begin{equation}
{\textit{L}({\Delta P_{t,i,d}}, \widetilde{\Delta P})} = \Sigma_{t=1}^{T} ( \alpha_1 || \Delta\textbf{R}_{t,i,d} - \widetilde{\Delta\textbf{R}_{t}} ||_2 + \alpha_2 ||\Delta\textbf{t}_{t,i,d} - \widetilde{\Delta\textbf{t}_{t}}||_2 )
\end{equation}

Besides the traditional pose loss, the state value loss is added. The state value has the ability to evaluate current pose prediction. To train the state value function, we regard the negative pose loss as ground truth $S_t$, and our state value prediction can be represented as $F(\Delta P_{i,d})$. So the state value pose ${\textit{L}}_{F(\Delta P_{t,i,d})}$ can be represented as following ($\alpha_3$ is the weight of state value loss): 

\begin{equation}
{\textit{L}}({F(\Delta P_{t,i,d})}, S_t) = \Sigma_{t=1}^{T} ( \alpha_3 || S_t - F(\Delta P_{t,i,d}) ||_2 )
\end{equation}

In conclusion, the total loss ${\textit{L}}_{oss}$ in our network is:
\begin{equation}
{\textit{L}}_{oss} = {\textit{L}({\Delta P_{t,i,d}}, \widetilde{\Delta P})} + {\textit{L}}({F(\Delta P_{t,i,d})}, S_t)
\end{equation}

\section{Experimental Results}
\label{sec:result}

\textbf{Agent in Data Augmentation}
To efficiently render depth image, we build an agent that encapsulates Open3D~\citep{Zhou2018} GL framework. Our agent contains the implementation of depth projection and depth image augmentation~\citep{bertalmio2000image}. To get more RGB and depth image pair, the pose of the camera is perturbed with random noise to render more depth images. Such perturbation results in image pairs with known ground truth $\Delta P$. In our experiments, the highest distance for two poses is about 12 m. In the indoor scenario, such a gap is considered very large because the view could be completely changed. The permutation rotation of the pose is $\pm 5^\circ$ and the translation permutation is $\pm 0.5m$. Every initial RGB and depth image pair can be augmented into 50 samples in our experiment. 
We use Lego-LOAM LiDAR SLAM \cite{shan2018lego} and sensor synchronization to get RGB keyframes. In total, we have more than $60000$ image pairs of keyframes with ground truth pose difference for parking garages $\{G_{ijk} | i\in[1,2], j\in[1,2,3], k\in[1,2]\}$. Each  dataset is divided into $60\%$ training, $30\%$ validation, and $10\%$ testing.

\textbf{Network Implementation}
Our network is implemented in Tensorflow 1.7.0, and the training process is on 4 GTX 1080 Titan GPU. The batch size of our model is setting to 16, and the training time for every batch is less than 1s. The optimizer we used here is RMSProp optimizer. In our experiments, we set the epoch number to 60. The learning rate is $10^{-3}$. In the training process, we will start train the pose net without the state value branch. When the model is more stable, the combining training is starting. As for the loss weight, we set $\alpha_1$, $\alpha_2$ and $\alpha_3$ to 100, 1, and 0.1 respectively. As for the inference time, every step is close 50 ms on Nvidia TX2 platform. 

\textbf{Pose/Localization Accuracy} 
We use the data from different parking garages at different times to demonstrate that our localization accuracy and generalizability.
Table~\ref{table:inference_setting} summarized some combinations of garage and time which we are using in the following experiments. 
\begin{table}
    \centering
    \begin{tabular}{c|c}
        \hline
        SPST & \textbf{s}ame \textbf{p}arking garage, \textbf{s}ame collection \textbf{t}ime \\ \hline
        SPDT & \textbf{s}ame \textbf{p}arking garage, \textbf{d}ifferent collection \textbf{t}ime \\ \hline
        SPDTDC & \textbf{s}ame \textbf{p}arking garage, \textbf{d}ifferent collection \textbf{t}ime, \textbf{d}ifferent \textbf{c}amera direction \\ \hline
        training-pure & the training and the inference data from same garage different trajectories \\ \hline
        training-mix & the training and the inference data from different garage different trajectories \\ \hline
    \end{tabular}
    \caption{Inference Setting Definitions}
    \label{table:inference_setting}
\end{table}

\emph{Performance within one garage}
The appearances of the same garage can be significantly different due to parking conditions and time of the day. To verify our model stays accurate when facing appearance changes, we experiment with the inference performance for several training settings with different RGB trajectories. 
Table~\ref{table:same_garage} contains the details about our localization accuracy for one garage. We can see the translation error is less than 10cm in the test data set on average cases, and the rotation error is close to 0 degree. The mixed training can get more stable results comparing to the pure training mode of SPDT, so more mix data can make our model more robust.  

\begin{table}
    \begin{tabular}{| l | l | l | l | l | l | l | l | l | l |}
    \hline 
    Source & Training & $E_{trans}$ & $E_{rotation}$ & $E_x$ & $E_y$ & $E_z$ & $E_{roll}$ & $E_{pitch}$ & $E_{yaw}$ \\ \hline \hline
    SPST & pure & 2.689E-2  & 3.035E-2 & 1.949E-2 & 7.772E-4 & 6.041E-2 & 9.123E-6 & 9.102E-2 & 1.334E-5 \\ \hline
    SPDT & pure & 1.077E1 & 6.467E-1 & 9.518 & 2.543E-3 & 2.279E1 & 2.156E-4 & 1.940 & 2.554E-4 \\ \hline
    SPDT & mix & 3.854E-2 & 6.494E-3 & 5.264E-2 & 1.111E-3 & 6.185E-2 & 1.245E-5 & 1.945E-2 & 1.614E-5 \\ \hline
    SPDTDC & pure & 2.436E-2 & 2.098E-2 & 1.732E-2 & 1.060E-3 & 5.470E-2 & 1.445E-4 & 6.256E2 & 2.432E-4 \\ \hline
    training-mix & mix & 2.436E-2 & 2.098E-2 & 1.732E-2 & 1.060E-3 & 5.470E-2 & 1.445E-4 & 6.256E2 & 2.432E-4 \\ \hline  
    \end{tabular}
    \caption{Same Garage, Different Training Settings}
    \label{table:same_garage}
\end{table}

\emph{Performance across garages}
We have just validated the model works well once being trained by the data from one single garage. What it requires for retraining on another garage is the key for generalization of the model. Therefore, we analyzed some fine-tuning schemes in the following. Table~\ref{table:diff_garage} indicates the mix version of training data will make the model faster to converge and get better result than pure version training. The pure version of training will make the training has large dependence on the order of sequence, which lowers the generalization ability of the model. Therefore, we have seen similar results in the same garage scenario and the cross-garage scenario based on similar intuition. Such results have good indications that more data in more parking garages makes the model more robust and accurate.

\begin{table}[h]
    \begin{tabular}{| l | l | l | l | l | l | l | l | l | l |}
    \hline 
    Source & Training  & $E_{trans}$ & $E_{rotation}$ & $E_x$ & $E_y$ & $E_z$ & $E_{roll}$ & $E_{pitch}$ & $E_{yaw}$ \\ \hline \hline
    DPDT & pure & 4.094 & 4.486E-2 & 1.126E1 & 7.497E-3 & 1.018 & 8.915E-4 & 1.328E-1 & 8.550E-4 \\ \hline
    DPDT & mix & 4.301E-2 & 2.246E-2 & 9.688E-2 & 4.429E-4 & 3.169E-2 & 1.232E-4 & 6.710E-2 & 1.409E-4  \\ \hline
    ALL & mix & 7.113E-2 & 7.918E-2 & 2.181E-1 & 5.526E-4 & 4.186 & 2.005E-5 & 1.452E-1 & 1.599E-5  \\ \hline
    \end{tabular}
    \caption{Different Garage, Different Training Settings}
    \label{table:diff_garage}
\end{table}

\begin{figure}[h]
    \centering
    \includegraphics[width=0.8\textwidth]{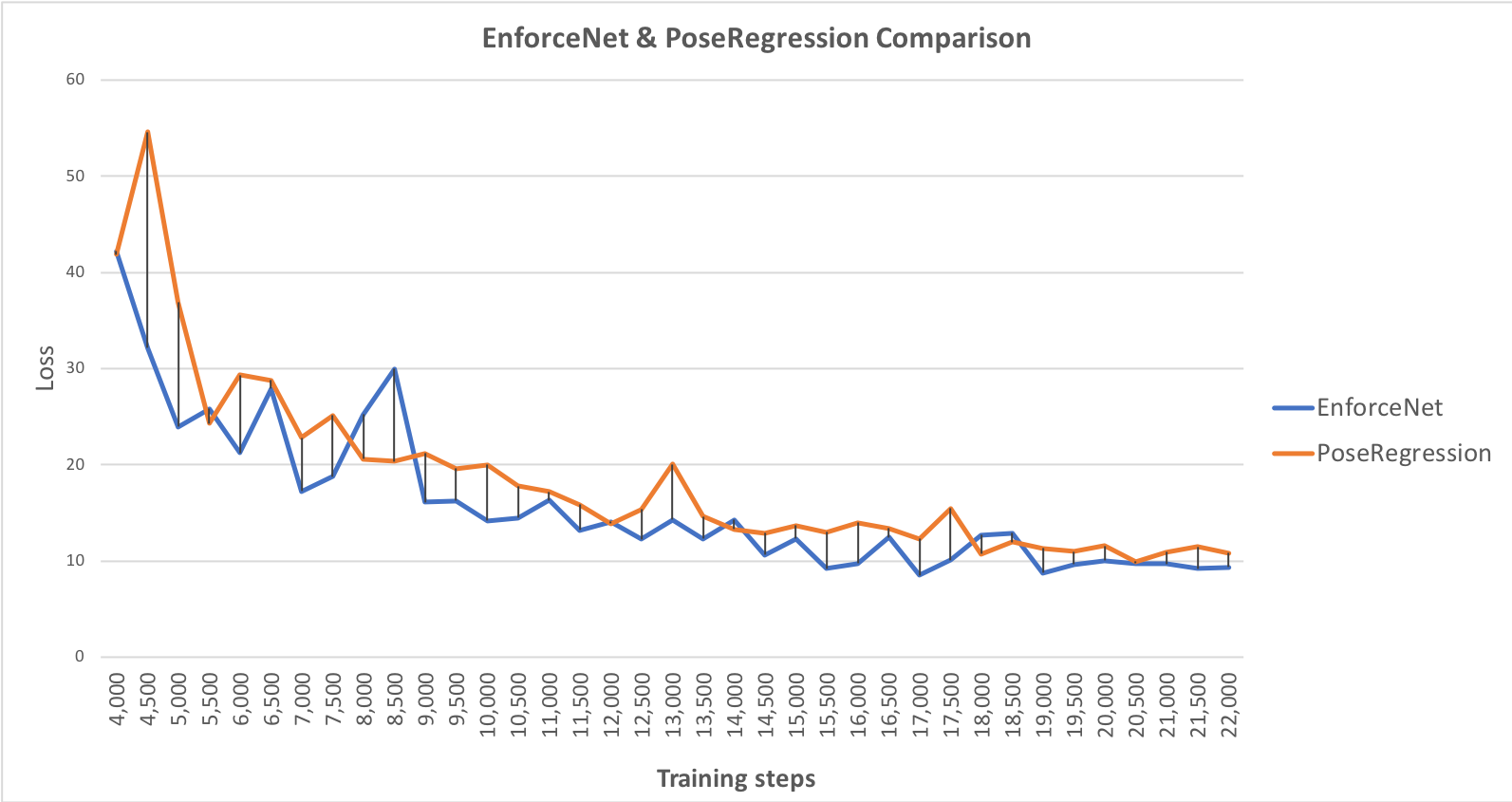}
    \caption{EnforceNet $\&$ PoseRegression Comparison}
    \label{fig:resistor}
\end{figure}

\emph{Performance of resistor module}
So far, we have demonstrated the weight-sharing and the EnforceNet structure works well for our task. We further show the performance difference between EnforceNet and a plain regression network on accuracy and generalizability. Figure~\ref{fig:resistor} compares the test pose loss with same training steps. From the figure, we can see the EnforceNet module can make our model converge faster and more stable, and improve the ability to a new scenario.

\section{Conclusion}
\label{sec:conclusion}

We proposed EnforceNet, an end-to-end solution for camera pose localization within a large scale and sparse 3D LiDAR point cloud. 
The EnforceNet has a novel resistor module and a weight-sharing scheme that is inspired by the state value function and value-iteration in RL framework. We conducted detailed experiments on real-world datasets of large scale indoor parking garages, and demonstrated the EnforceNet has reached the state-of-the-art localization accuracy with better generalization capability than previous research. We are preparing to open-source our system and the dataset for the ease of further development of similar research for the community. 
	


\clearpage


\bibliography{example}  

\begin{thebibliography}{37}
\providecommand{\natexlab}[1]{#1}
\providecommand{\url}[1]{\texttt{#1}}
\expandafter\ifx\csname urlstyle\endcsname\relax
  \providecommand{\doi}[1]{doi: #1}\else
  \providecommand{\doi}{doi: \begingroup \urlstyle{rm}\Url}\fi

\bibitem[Levinson and Thrun(2010)]{sebastian07}
J.~Levinson and S.~Thrun.
\newblock Robust vehicle localization in urban environments using probabilistic
  maps.
\newblock In \emph{Proceedings of IEEE International Conference on Robotics and
  Automation}, 2010.

\bibitem[Forster et~al.(2014)Forster, Pizzoli, and Scaramuzza]{forster2014svo}
C.~Forster, M.~Pizzoli, and D.~Scaramuzza.
\newblock Svo: Fast semi-direct monocular visual odometry.
\newblock In \emph{2014 IEEE international conference on robotics and
  automation (ICRA)}, pages 15--22. IEEE, 2014.

\bibitem[Engel et~al.(2017)Engel, Koltun, and Cremers]{engel2017direct}
J.~Engel, V.~Koltun, and D.~Cremers.
\newblock Direct sparse odometry.
\newblock \emph{IEEE transactions on pattern analysis and machine
  intelligence}, 40\penalty0 (3):\penalty0 611--625, 2017.

\bibitem[Davison et~al.(2007)Davison, Reid, Molton, and
  Stasse]{davison2007monoslam}
A.~J. Davison, I.~D. Reid, N.~D. Molton, and O.~Stasse.
\newblock Monoslam: Real-time single camera slam.
\newblock \emph{IEEE Transactions on Pattern Analysis \& Machine Intelligence},
  \penalty0 (6):\penalty0 1052--1067, 2007.

\bibitem[Klein and Murray(2009)]{klein2009parallel}
G.~Klein and D.~Murray.
\newblock Parallel tracking and mapping on a camera phone.
\newblock In \emph{2009 8th IEEE International Symposium on Mixed and Augmented
  Reality}, pages 83--86. IEEE, 2009.

\bibitem[Engel et~al.(2014)Engel, Sch{\"o}ps, and Cremers]{engel2014lsd}
J.~Engel, T.~Sch{\"o}ps, and D.~Cremers.
\newblock Lsd-slam: Large-scale direct monocular slam.
\newblock In \emph{European conference on computer vision}, pages 834--849.
  Springer, 2014.

\bibitem[Mur-Artal and Tard{\'o}s(2017)]{mur2017orb}
R.~Mur-Artal and J.~D. Tard{\'o}s.
\newblock Orb-slam2: An open-source slam system for monocular, stereo, and
  rgb-d cameras.
\newblock \emph{IEEE Transactions on Robotics}, 33\penalty0 (5):\penalty0
  1255--1262, 2017.

\bibitem[Cadena et~al.(2016)Cadena, Carlone, Carrillo, Latif, Scaramuzza,
  Neira, Reid, and Leonard]{cadena2016past}
C.~Cadena, L.~Carlone, H.~Carrillo, Y.~Latif, D.~Scaramuzza, J.~Neira, I.~Reid,
  and J.~J. Leonard.
\newblock Past, present, and future of simultaneous localization and mapping:
  Toward the robust-perception age.
\newblock \emph{IEEE Transactions on robotics}, 32\penalty0 (6):\penalty0
  1309--1332, 2016.

\bibitem[Bergmann et~al.(2017)Bergmann, Wang, and Cremers]{bergmann2017online}
P.~Bergmann, R.~Wang, and D.~Cremers.
\newblock Online photometric calibration of auto exposure video for realtime
  visual odometry and slam.
\newblock \emph{IEEE Robotics and Automation Letters}, 3\penalty0 (2):\penalty0
  627--634, 2017.

\bibitem[Geiger et~al.(2012)Geiger, Lenz, and Urtasun]{Geiger2012CVPR}
A.~Geiger, P.~Lenz, and R.~Urtasun.
\newblock Are we ready for autonomous driving? the kitti vision benchmark
  suite.
\newblock In \emph{Conference on Computer Vision and Pattern Recognition
  (CVPR)}, 2012.

\bibitem[Sutton and Barto(2018)]{sutton2018reinforcement}
R.~S. Sutton and A.~G. Barto.
\newblock \emph{Reinforcement learning: An introduction}.
\newblock MIT press, 2018.

\bibitem[Caselitz et~al.(2016{\natexlab{a}})Caselitz, Steder, Ruhnke, and
  Burgard]{caselitz2016monocular}
T.~Caselitz, B.~Steder, M.~Ruhnke, and W.~Burgard.
\newblock Monocular camera localization in 3d lidar maps.
\newblock In \emph{2016 IEEE/RSJ International Conference on Intelligent Robots
  and Systems (IROS)}, pages 1926--1931. IEEE, 2016{\natexlab{a}}.

\bibitem[Caselitz et~al.(2016{\natexlab{b}})Caselitz, Steder, Ruhnke, and
  Burgard]{caselitz2016matching}
T.~Caselitz, B.~Steder, M.~Ruhnke, and W.~Burgard.
\newblock Matching geometry for long-term monocular camera localization.
\newblock In \emph{ICRA Workshop: AI for long-term Autonomy},
  2016{\natexlab{b}}.

\bibitem[Gawel et~al.(2016)Gawel, Cieslewski, Dub{\'e}, Bosse, Siegwart, and
  Nieto]{gawel2016structure}
A.~Gawel, T.~Cieslewski, R.~Dub{\'e}, M.~Bosse, R.~Siegwart, and J.~Nieto.
\newblock Structure-based vision-laser matching.
\newblock In \emph{2016 IEEE/RSJ International Conference on Intelligent Robots
  and Systems (IROS)}, pages 182--188. IEEE, 2016.

\bibitem[Saurer et~al.(2016)Saurer, Baatz, K{\"o}ser, Pollefeys,
  et~al.]{saurer2016image}
O.~Saurer, G.~Baatz, K.~K{\"o}ser, M.~Pollefeys, et~al.
\newblock Image based geo-localization in the alps.
\newblock \emph{International Journal of Computer Vision}, 116\penalty0
  (3):\penalty0 213--225, 2016.

\bibitem[Pandey et~al.(2015)Pandey, McBride, Savarese, and
  Eustice]{pandey2015automatic}
G.~Pandey, J.~R. McBride, S.~Savarese, and R.~M. Eustice.
\newblock Automatic extrinsic calibration of vision and lidar by maximizing
  mutual information.
\newblock \emph{Journal of Field Robotics}, 32\penalty0 (5):\penalty0 696--722,
  2015.

\bibitem[Napier et~al.(2013)Napier, Corke, and Newman]{napier2013cross}
A.~Napier, P.~Corke, and P.~Newman.
\newblock Cross-calibration of push-broom 2d lidars and cameras in natural
  scenes.
\newblock In \emph{2013 IEEE International Conference on Robotics and
  Automation}, pages 3679--3684. IEEE, 2013.

\bibitem[Wolcott and Eustice(2014)]{wolcott2014visual}
R.~W. Wolcott and R.~M. Eustice.
\newblock Visual localization within lidar maps for automated urban driving.
\newblock In \emph{2014 IEEE/RSJ International Conference on Intelligent Robots
  and Systems}, pages 176--183. IEEE, 2014.

\bibitem[Pascoe et~al.(2015)Pascoe, Maddern, and Newman]{pascoe2015direct}
G.~Pascoe, W.~Maddern, and P.~Newman.
\newblock Direct visual localisation and calibration for road vehicles in
  changing city environments.
\newblock In \emph{Proceedings of the IEEE International Conference on Computer
  Vision Workshops}, pages 9--16, 2015.

\bibitem[Neubert et~al.(2017)Neubert, Schubert, and
  Protzel]{neubert2017sampling}
P.~Neubert, S.~Schubert, and P.~Protzel.
\newblock Sampling-based methods for visual navigation in 3d maps by
  synthesizing depth images.
\newblock In \emph{2017 IEEE/RSJ International Conference on Intelligent Robots
  and Systems (IROS)}, pages 2492--2498. IEEE, 2017.

\bibitem[Caron et~al.(2014)Caron, Dame, and Marchand]{caron2014direct}
G.~Caron, A.~Dame, and E.~Marchand.
\newblock Direct model based visual tracking and pose estimation using mutual
  information.
\newblock \emph{Image and Vision Computing}, 32\penalty0 (1):\penalty0 54--63,
  2014.

\bibitem[Naseer and Burgard(2017)]{naseer2017deep}
T.~Naseer and W.~Burgard.
\newblock Deep regression for monocular camera-based 6-dof global localization
  in outdoor environments.
\newblock In \emph{2017 IEEE/RSJ International Conference on Intelligent Robots
  and Systems (IROS)}, pages 1525--1530. IEEE, 2017.

\bibitem[Wang et~al.(2018)Wang, Yang, Cao, Xu, and Lin]{wang2018dels}
P.~Wang, R.~Yang, B.~Cao, W.~Xu, and Y.~Lin.
\newblock Dels-3d: Deep localization and segmentation with a 3d semantic map.
\newblock In \emph{Proceedings of the IEEE Conference on Computer Vision and
  Pattern Recognition}, pages 5860--5869, 2018.

\bibitem[Radwan et~al.(2018)Radwan, Valada, and Burgard]{radwan2018vlocnet++}
N.~Radwan, A.~Valada, and W.~Burgard.
\newblock Vlocnet++: Deep multitask learning for semantic visual localization
  and odometry.
\newblock \emph{IEEE Robotics and Automation Letters}, 3\penalty0 (4):\penalty0
  4407--4414, 2018.

\bibitem[Wang et~al.(2019)Wang, Xu, Zhu, Mart{\'\i}n-Mart{\'\i}n, Lu, Fei-Fei,
  and Savarese]{wang2019densefusion}
C.~Wang, D.~Xu, Y.~Zhu, R.~Mart{\'\i}n-Mart{\'\i}n, C.~Lu, L.~Fei-Fei, and
  S.~Savarese.
\newblock Densefusion: 6d object pose estimation by iterative dense fusion.
\newblock \emph{arXiv preprint arXiv:1901.04780}, 2019.

\bibitem[Triggs et~al.(1999)Triggs, McLauchlan, Hartley, and
  Fitzgibbon]{triggs1999bundle}
B.~Triggs, P.~F. McLauchlan, R.~I. Hartley, and A.~W. Fitzgibbon.
\newblock Bundle adjustment—a modern synthesis.
\newblock In \emph{International workshop on vision algorithms}, pages
  298--372. Springer, 1999.

\bibitem[Kendall et~al.(2015)Kendall, Grimes, and Cipolla]{kendall2015posenet}
A.~Kendall, M.~Grimes, and R.~Cipolla.
\newblock Posenet: A convolutional network for real-time 6-dof camera
  relocalization.
\newblock In \emph{Proceedings of the IEEE international conference on computer
  vision}, pages 2938--2946, 2015.

\bibitem[Zhou et~al.(2017)Zhou, Brown, Snavely, and Lowe]{zhou2017unsupervised}
T.~Zhou, M.~Brown, N.~Snavely, and D.~G. Lowe.
\newblock Unsupervised learning of depth and ego-motion from video.
\newblock In \emph{Proceedings of the IEEE Conference on Computer Vision and
  Pattern Recognition}, pages 1851--1858, 2017.

\bibitem[Li et~al.(2018)Li, Wang, Long, and Gu]{li2018undeepvo}
R.~Li, S.~Wang, Z.~Long, and D.~Gu.
\newblock Undeepvo: Monocular visual odometry through unsupervised deep
  learning.
\newblock In \emph{2018 IEEE International Conference on Robotics and
  Automation (ICRA)}, pages 7286--7291. IEEE, 2018.

\bibitem[Mahjourian et~al.(2018)Mahjourian, Wicke, and
  Angelova]{mahjourian2018unsupervised}
R.~Mahjourian, M.~Wicke, and A.~Angelova.
\newblock Unsupervised learning of depth and ego-motion from monocular video
  using 3d geometric constraints.
\newblock In \emph{Proceedings of the IEEE Conference on Computer Vision and
  Pattern Recognition}, pages 5667--5675, 2018.

\bibitem[Casser et~al.(2019)Casser, Pirk, Mahjourian, and
  Angelova]{casser2019unsupervised}
V.~Casser, S.~Pirk, R.~Mahjourian, and A.~Angelova.
\newblock Unsupervised monocular depth and ego-motion learning with structure
  and semantics.
\newblock In \emph{Proceedings of the IEEE Conference on Computer Vision and
  Pattern Recognition Workshops}, pages 0--0, 2019.

\bibitem[Mirowski et~al.(2018)Mirowski, Grimes, Malinowski, Hermann, Anderson,
  Teplyashin, Simonyan, Zisserman, Hadsell, et~al.]{mirowski2018learning}
P.~Mirowski, M.~Grimes, M.~Malinowski, K.~M. Hermann, K.~Anderson,
  D.~Teplyashin, K.~Simonyan, A.~Zisserman, R.~Hadsell, et~al.
\newblock Learning to navigate in cities without a map.
\newblock In \emph{Advances in Neural Information Processing Systems}, pages
  2419--2430, 2018.

\bibitem[Kato et~al.(2018)Kato, Tokunaga, Maruyama, Maeda, Hirabayashi,
  Kitsukawa, Monrroy, Ando, Fujii, and Azumi]{kato2018autoware}
S.~Kato, S.~Tokunaga, Y.~Maruyama, S.~Maeda, M.~Hirabayashi, Y.~Kitsukawa,
  A.~Monrroy, T.~Ando, Y.~Fujii, and T.~Azumi.
\newblock Autoware on board: Enabling autonomous vehicles with embedded
  systems.
\newblock In \emph{2018 ACM/IEEE 9th International Conference on Cyber-Physical
  Systems (ICCPS)}, pages 287--296. IEEE, 2018.

\bibitem[Shan and Englot(2018)]{shan2018lego}
T.~Shan and B.~Englot.
\newblock Lego-loam: Lightweight and ground-optimized lidar odometry and
  mapping on variable terrain.
\newblock In \emph{2018 IEEE/RSJ International Conference on Intelligent Robots
  and Systems (IROS)}, pages 4758--4765. IEEE, 2018.

\bibitem[Pham et~al.(2018)Pham, Guan, Zoph, Le, and Dean]{pham2018efficient}
H.~Pham, M.~Y. Guan, B.~Zoph, Q.~V. Le, and J.~Dean.
\newblock Efficient neural architecture search via parameter sharing.
\newblock \emph{arXiv preprint arXiv:1802.03268}, 2018.

\bibitem[Zhou et~al.(2018)Zhou, Park, and Koltun]{Zhou2018}
Q.-Y. Zhou, J.~Park, and V.~Koltun.
\newblock {Open3D}: {A} modern library for {3D} data processing.
\newblock \emph{arXiv:1801.09847}, 2018.

\bibitem[Bertalmio et~al.(2000)Bertalmio, Sapiro, Caselles, and
  Ballester]{bertalmio2000image}
M.~Bertalmio, G.~Sapiro, V.~Caselles, and C.~Ballester.
\newblock Image inpainting.
\newblock In \emph{Proceedings of the 27th annual conference on Computer
  graphics and interactive techniques}, pages 417--424. ACM
  Press/Addison-Wesley Publishing Co., 2000.

\end{thebibliography}

\end{document}